\RequirePackage{fix-cm}
\documentclass[smallextended,a4paper,final]{svjour3}       
\smartqed  
\usepackage{makeidx,amsmath,epsfig,url,multirow,array,color,multirow,bigstrut}  %
\usepackage{hyperref}
\usepackage{graphicx, listings, xcolor}
\usepackage[caption=false]{subfig}

\usepackage[ruled,linesnumbered,lined,boxed,commentsnumbered]{algorithm2e}

\usepackage[nocompress]{cite}

\usepackage{geometry}
\usepackage{array}
\newcolumntype{L}[1]{>{\raggedright\let\newline\\\arraybackslash\hspace{0pt}}m{#1}}
\newcolumntype{C}[1]{>{\centering\let\newline\\\arraybackslash\hspace{0pt}}m{#1}}
\newcolumntype{R}[1]{>{\raggedleft\let\newline\\\arraybackslash\hspace{0pt}}m{#1}}

\def\etal{\emph{et al.}~}
\def\ie{\emph{i.e.}~}
\newcommand{\thickhline}{%
    \noalign {\ifnum 0=`}\fi \hrule height 1pt
    \futurelet \reserved@a \@xhline
}
\newcolumntype{"}{@{\hskip\tabcolsep\vrule width 1pt\hskip\tabcolsep}}

\newcommand{\rev}[1]{#1}

\journalname{Multimedia Tools and Applications}
\begin{document}

\title{M-VAD Names:\\a Dataset for Video Captioning with Naming}

\subtitle{\small{\url{https://github.com/aimagelab/mvad-names-dataset}}}

\titlerunning{Pini~\emph{et al.}: M-VAD Names: a Dataset for Video Captioning with Naming}        

\author{Stefano Pini \and
        Marcella Cornia \and~~~~~
        Federico Bolelli \and
        Lorenzo Baraldi \and~~~~~
        Rita Cucchiara
}

\authorrunning{\textit{Multimedia Tools and Applications}, 2018} 

\institute{S. Pini, M. Cornia, F. Bolelli, L. Baraldi, and R. Cucchiara \at
           Department of Engineering ``Enzo Ferrari'' \\
           University of Modena and Reggio Emilia \\
           \email{$\{$s.pini,marcella.cornia,federico.bolelli,lorenzo.baraldi,rita.cucchiara$\}$@unimore.it}           
}

\date{Received: February 2018 / Accepted: December 2018}

\maketitle

\begin{abstract}
Current movie captioning architectures are not capable of mentioning characters with their proper name, replacing them with a generic ``someone'' tag. \rev{The lack of movie description datasets with characters' visual annotations surely plays a relevant role in this shortage.}
\rev{Recently, we proposed to extend the M-VAD dataset by introducing such information. In this paper, we present an improved version of the dataset, namely M-VAD Names, and its semi-automatic annotation procedure. The resulting dataset contains $63$k visual tracks and $34$k textual mentions, all associated with character identities.
To showcase the features of the dataset and quantify the complexity of the naming task, we investigate multimodal architectures to replace the ``someone'' tags with proper character names in existing video captions. The evaluation is further extended by testing this application on videos outside of the M-VAD Names dataset.}

\end{abstract}

\section{Introduction}
In the past few years, video captioning has gained more and more attention, thanks to the release of large-scale movie description datasets~\cite{torabi2015using,rohrbach15cvpr} and the development of deep learning-based algorithms~\cite{venugopalan2015sequence,pan2015jointly}. Nowadays, the research is moving its attention to the quality of the captions~\cite{shetty2017speaking} and to the inclusion of the visual semantic information~\cite{rohrbach2015long}. \rev{Nonetheless, an important feature which is still missing in movie captioning models is the ability to mention characters with their proper names.}

\rev{As a matter of fact, it is also a common practice to replace character names with a generic ``someone'' tag when building novel datasets.}
The underlying reason has to be found in the structure of current video captioning models, which are not designed to take into account the visual aspect of each movie character. In these architectures, using captions in which character names are not replaced would simply result in additional dictionary entries, ignoring the fundamental relationship between the character names and their visual appearance, \rev{and possibly invalidating the significance of evaluation metrics}.

Developing video captioning architectures with naming capabilities requires to deal with several sub-tasks at the time of caption generation. In particular, the architecture has to detect, track, and recognize people within a set of characters. \rev{Furthermore, the language model has to be aware of the semantic structure of the caption and has to coordinate itself with the feature extraction part, to detect the presence of a character in the scene.}

Unfortunately, current movie description datasets do not contain any kind of supervision that joins the textual mentions and the visual appearances of the characters. Without a supervision between the textual and the visual domain, training video captioning algorithms with naming capabilities is particularly challenging. Indeed, many characters and background actors may appear in the same scene, while only few characters are mentioned in the video descriptions. Therefore, an additional form of supervision that associates the textual and the visual information, by linking characters' visual appearances with their textual mentions, is necessary for the development of novel movie description architectures.

In this paper, we introduce a novel version of the M-VAD Names dataset, specifically designed for supporting the development of video captioning architectures with naming capabilities. The dataset, which is an extension of the well-known Montreal Video Description Dataset (M-VAD), consists of visual face tracks and the association between them and the characters' textual mentions.
\rev{With respect to the previous version of the dataset presented in~\cite{pini2017towards}, this release introduces different improvements, which include longer and more precise face tracks, a refined version of the semi-automatic annotation procedure, the manual correction of errors in the original M-VAD captions, and the manual correction of errors in the proposed annotations. Therefore, the resulting dataset is more precise in the annotation and more useful in the data it contains. In the rest of the paper, we discuss those improvements in detail and compare with respect to the old release of the dataset.}

In addition, we propose a multimodal architecture that addresses the task of replacing generic ``someone'' tags with proper character names in previously generated captions. The model combines advanced Natural Language Processing tools and state-of-the-art deep neural models for action and face recognition. Experimental results enlighten and quantify many of the challenges associated with the task, and demonstrate the effectiveness of the proposed strategy.
\rev{Finally, we also show how the proposed model can be applied outside of the M-VAD Names dataset, by extending the evaluation on an additional set of movies.}

\section{Related work}
\label{sec:related}
Our work is related to the task of linking visual tracks in the context of movies and TV series to the proper character names, and to the generation of captions for video. In this section, we review related work in these two directions.

\subsection{Linking visual tracks to names}
The problem of identifying characters in movies or TV series has been widely addressed by computer vision researchers who principally focus on linking people with their names by tracking faces in the video and assigning names to them~\cite{bojanowski2013finding,everingham2006hello,ramanathan2014linking,sivic2009you,tapaswi2012knock}. For example,~\cite{everingham2006hello,sivic2009you} tackled this problem by automatically aligning subtitles and script texts of movies and TV series. In particular, Everingham~\etal\cite{everingham2006hello} aimed to associate speaker names present in the movie scripts to the correct faces appearing in the movie clips by detecting face tracks with lip motion. 
Sivic~\etal\cite{sivic2009you} extended the previous work, limited in classifying frontal faces, by adding the detection and recognition of characters in profile views, improving the overall performance.

In~\cite{tapaswi2012knock}, each TV series episode is instead modelled as a Markov Random Field, integrating cues from face, speech, and clothing. Bojanowski~\etal\cite{bojanowski2013finding} proposed a method to extract actor/action pairs from movie scripts and used them as constraints in a discriminative clustering framework. In~\cite{ramanathan2014linking}, authors introduced a joint model for person naming and co-reference resolution which consists in resolving the identity of ambiguous mentions of people such as pronouns (e.g. ``he'' or ``she'') and nominals (e.g. ``man'').

Recently, Rohrbach~\etal\cite{RohrbachCVPR2017a} addressed the problem of generating video descriptions with grounded and co-referenced people by proposing a deeply-learned model. This task significantly differs from the one tackled in this paper, as it aims at predicting the spatial location in which a given character appears, and at producing captions with proper names in the correct place.

Miech~\etal\cite{miech2017learning}, instead, addressed the problem of weakly supervised learning of actions and actors from movies by applying an online optimization algorithm based on the Block-Coordinate Frank-Wolfe method. Finally, in~\cite{jin2017end} an end-to-end system for detecting and clustering faces by identity in full-length movies is proposed. However, this approach is far from the aforementioned works as it only aimed at clustering face tracks without naming the corresponding movie characters.

Several other methods have been proposed towards understanding social aspects of movies and TV series scenes for either classifying different types of interactions~\cite{patron2012structured} or predicting whether people are looking at each other~\cite{ding2010learning,marin2014detecting}. As an example, Vicol~\etal\cite{moviegraphs2017} introduced a novel dataset which provides graph-based annotations of social situations appearing in movie clips to capture who is present in the clip, their emotional and physical attributes, their relationships, and the interactions between them. 

\subsection{Video captioning}
The generation of natural language descriptions of 
visual content has received large interest since the emergence of recurrent networks, either for single images~\cite{hendricks2015deep}, user-generated videos~\cite{donahue2015long}, or movie clips~\cite{rohrbach2015long,venugopalan2015sequence}. First approaches described the input video through mean-pooled CNN features~\cite{venugopalan2014translating} or sequentially encoded by a recurrent layer~\cite{donahue2015long,venugopalan2015sequence}.
This strategy was then followed by the majority of video captioning approaches, either by incorporating attentive mechanisms~\cite{yao2015describing} in the sentence decoder, by building a common visual-semantic embedding~\cite{pan2015jointly}, or by adding external knowledge with language models~\cite{venugopalan16emnlp} or visual classifiers~\cite{rohrbach2015long}. 

Recent video captioning models have improved both components of the encoder-decoder approach by significantly changing their structure. Yu~\etal\cite{yu2016video} focused on the sentence decoder and proposed a hierarchical model containing a sentence and a paragraph generator. In particular, the sentence generator produces one simple short sentence that describes a specific short video interval by exploiting both temporal and spatial attention mechanisms. In contrast, Pan~\etal\cite{pan2016hierarchical} concentrated on the video encoding stage and introduced a hierarchical recurrent encoder to exploit temporal information of videos. In~\cite{baraldi2017hierarchical}, instead, authors proposed a modification of the LSTM cell able to identify discontinuity points between frames or segments and to modify the temporal connections of the encoding layer accordingly.

On a different note, Krishna~\etal\cite{krishna2017dense} introduced the task of dense-captioning events, which involves both detecting and describing events in a
video, and proposed a new model able to identify events of a video while simultaneously describing the detected events in natural language.

We believe that our dataset could be used in the video captioning domain moving researches towards the generation of video captioning architectures with naming capabilities.

\section{The M-VAD Names dataset}
\begin{figure}[t!]
    \centering
    \renewcommand{\arraystretch}{1.5}
    \begin{tabular}{c}
        \includegraphics[width=0.96\textwidth]{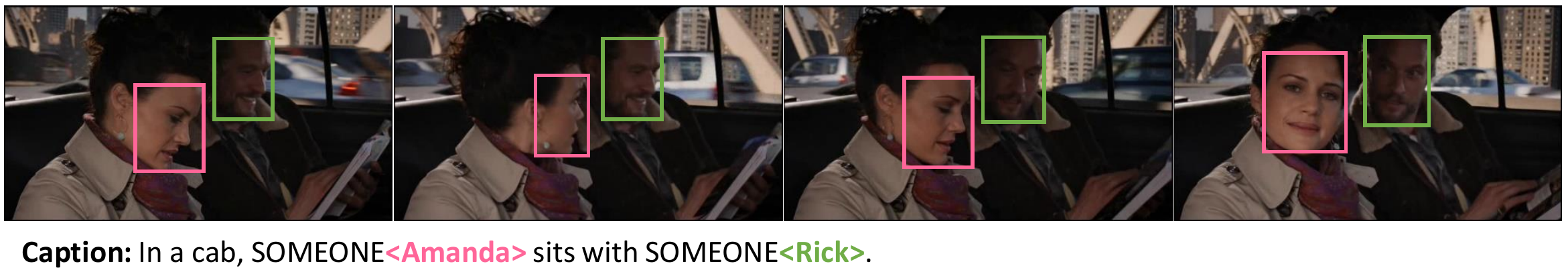}\\
        \includegraphics[width=0.96\textwidth]{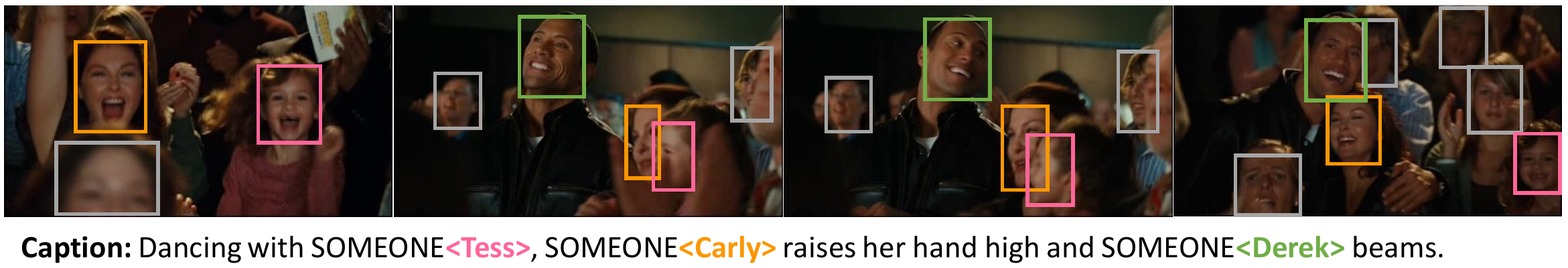}\\
        \includegraphics[width=0.96\textwidth]{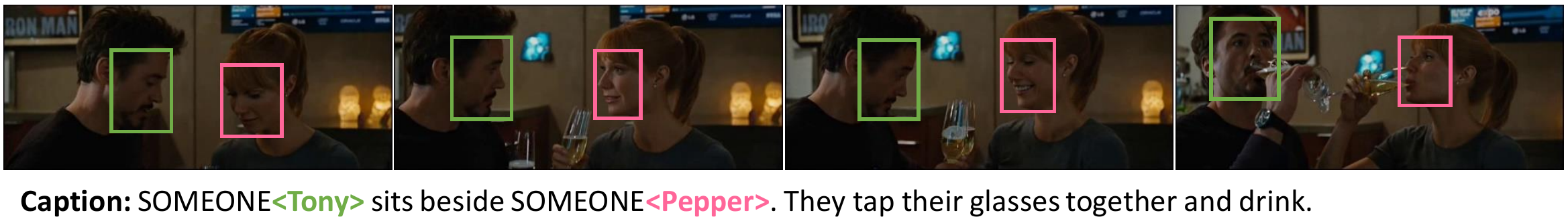}\\
        \includegraphics[width=0.96\textwidth]{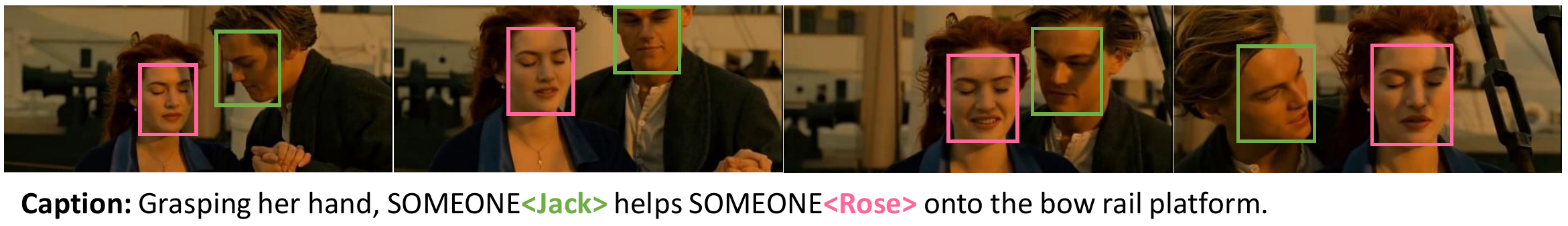}\\
        \includegraphics[width=0.96\textwidth]{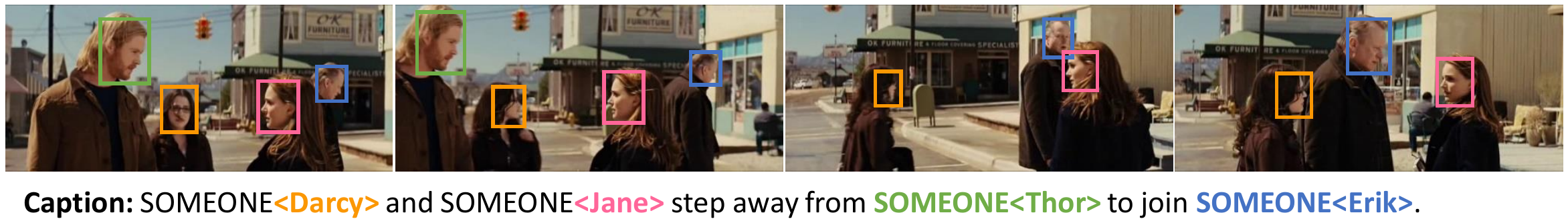}\\
        \includegraphics[width=0.96\textwidth]{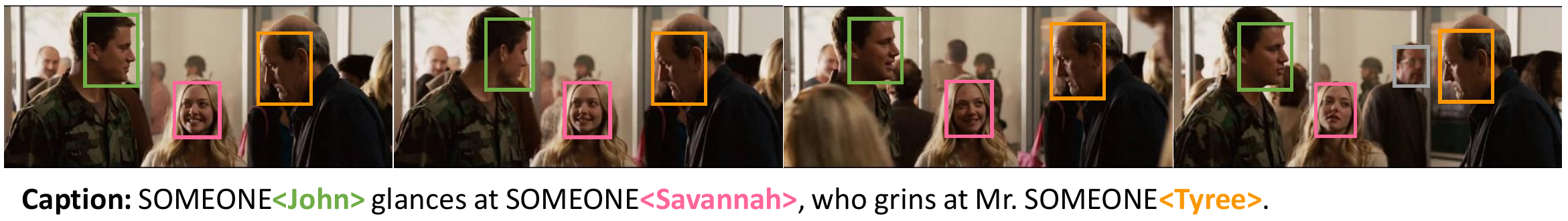}\\
        \includegraphics[width=0.96\textwidth]{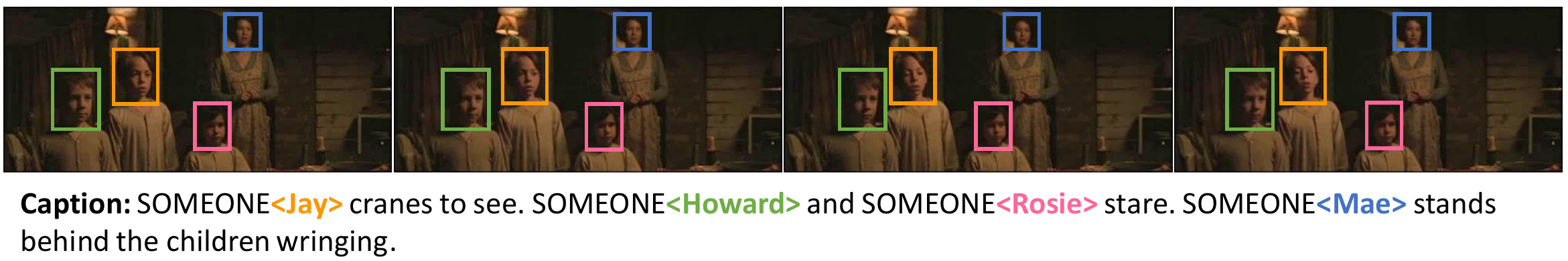}\\
    \end{tabular}
    \caption{Samples extracted from the M-VAD Names dataset. For each video clip, face tracks are annotated and associated to proper character names mentioned inside the caption. Face tracks that are not associated to a specific movie character (\ie unknown people) are represented in gray color.}
    \label{fig:samples}
\end{figure}
We collect and release a refined and extended version of the M-VAD Names dataset\footnote{\rev{The proposed dataset is publicly available at \url{https://github.com/aimagelab/mvad-names-dataset}.}}, a new set of annotations for the Montreal Video Annotation Dataset (M-VAD)~\cite{torabi2015using} supporting the development of video captioning architectures with naming capabilities.
The dataset contains the annotations of the characters' visual appearances, in the form of tracks of face bounding boxes, and the associations with the characters' textual mentions, when available.
\rev{In particular, we detect and annotate the visual appearances of characters in each video clip of each movie through a semi-automatic approach which extends~\cite{pini2017towards,jin2017end}.} Also, we correct some errors in the original M-VAD annotations in order to include more characters in our dataset. Figure~\ref{fig:samples} shows some representative samples of the collected dataset.

\rev{With respect to the first version that we presented in~\cite{pini2017towards}, the new M-VAD Names dataset contains face tracks only, since body tracks directly depend on the corresponding face tracks. The semi-automatic annotation procedure has been revisited to create more precise face tracks which are no more limited to a fixed size of $16$ frames, and to deal with characters which have multiple names within the movie. Finally, all the annotations of both the original M-VAD dataset and our annotation procedure have been manually corrected.}

In this section, we describe the refined annotation procedure, from the detection of the face tracks to the semi-automatic annotation process, the generation of the train, validation, and test split, and the method used to extend the original M-VAD captions. Finally, we report statistics and analyses of the proposed dataset, highlighting the improvements with respect to the previous version of M-VAD Names.

\subsection{Face detection and tracking}
The first stage of the annotation procedure is the extraction of the face tracks, sequences of consecutive face detections belonging to the same character.

To collect them, we sequentially detect faces in each frame of each video clip using the face detector presented in~\cite{ZhangZL016}. Then, tracks are formed by grouping consecutive detections belonging to the same character. Specifically, for each detected face, a tracker~\cite{Babenko09visualtracking} is initialized with the bounding box corresponding to the detection and a new face track is created.
Then, in the following frames, each initialized tracker is updated and each face detection is compared with each tracker prediction using the Intersection over Union (IoU) measure.
Applying the Kuhn-Munkres algorithm~\cite{kuhn1955hungarian}, each face detection is associated with the most overlapping tracker prediction.
Then, if the IoU value between a face detection and the associated tracker prediction is over a threshold (called $t_{IoU}$) and the appearance difference between the detection and the last element of the track is below a threshold (called $t_{visual}$), the face detection is added to the
related track
and the tracker is re-initialized on the new detection.
Otherwise,
a new tracker and a new track are initialized.
We empirically set the IoU threshold value to $0.5$, while
we found that a pixel-wise difference between face detections and last added element of tracks is, when used with a threshold of $10$, a sufficient appearance measure to discard most of the errors.

If a tracker prediction is not associated with any face detection (due to occlusions or scene changes, for instance) or the association does not respect the constraints reported above, the tracker prediction is added to the track. If the tracker is not associated again with a face detection for the following $8$ frames, or before the end of the video clip, the tracker predictions that were added to the track are removed and the tracker is detached.

At the end of each video clip, face tracks that are composed by less than $8$ frames are discarded, as well as tracks that are fully contained in another one.

The overall algorithm for the detection and tracking of face tracks is reported in Algorithm~\ref{alg:tracklets}.

\begin{algorithm}[t]
 \label{alg:tracklets}
 \begin{small}
 \SetInd{0.2em}{0.7em}
 \KwData{M-VAD dataset}
 \KwResult{Tracks containing characters' visual appearances}
 \ForEach{video clip {\bf in} M-VAD dataset}{
   \ForEach{frame {\bf in} video clip}{
    \lForEach{initialized tracker}{Update the tracker prediction}
    Detect faces in the frame (MTCNN architecture)\;
    Calc the IoU between each prediction and each detection\;
    Solve the detection-tracker association (Kuhn-Munkres algorithm)\;
    \ForEach{face detection which is not associated}{
     Create a new track with the face detection\;
     Define a new tracker linked to the new track\;
     Initialize the tracker on the face detection\;
    }
    \ForEach{initialized tracker}{
     \eIf{tracker is associated {\bf and} $IoU > t_{IoU}$ {\bf and} visual difference $ < t_{visual}$}{
      Add the detection to the track associated with the tracker\;
      Re-initialize the tracker with the new face detection\;
      Reset the tracker counter\;
     }
     {
      \eIf{$tracker_{counter} < t_{counter}$}{
       Add the tracker prediction to the associated track\;
       Increment the tracker counter\;
      }
      {
       Remove the last $tracker_{counter}$ items from the track\;
       Detach the tracker\;
      }
     }
    }
   }
   \ForEach{initialized tracker with $tracker_{counter} > 0$}{
     Remove the last $tracker_{counter}$ items from the track\;
   }
 }
 \end{small}
\caption{Track extraction algorithm.}
\end{algorithm}

\subsection{Movie character annotations}\label{sec:mca}
After the extraction, each face track has to be labelled with the name of a character from the corresponding movie. \rev{To facilitate the annotation procedure, which would require to label every single track with respect to the list of characters of a movie, we firstly cluster similar faces using an embedding space in which similar faces (\ie faces of the same person) lie together, while dissimilar faces (\ie faces of different people) lie far by a clear margin. Then, clusters are manually verified, to guarantee that each cluster contains only tracks from a single character. The annotator is finally asked to match each cluster with the corresponding character.}

To obtain the embedding space, we extract face feature vectors using a deep neural model inspired by FaceNet~\cite{schroff2015facenet} and trained on a sub-set of the MS-Celeb-1M dataset~\cite{guo2016ms}.
Then, for each movie, we aggregate tracks containing similar faces by applying a hierarchical clustering algorithm, based on the euclidean distance and on the Ward's minimum variance method~\cite{ward1963hierarchical}. Since each track is composed by a variable number of bounding boxes, we apply the clustering algorithm on the mean of their embeddings. We exclude the smallest tracks (\ie tracks with a side lower than $28$ pixels) from the automatic clustering process as we found that their features are not reliable.

\rev{Once clusters have been manually verified, so to contain one single character, each cluster is either assigned to a character of the movie, or rejected as \textit{wrong} (if it contains false positive detections by the face detector or by the tracker), or as \textit{unknown} (if the character is a background actor or if human annotators are not able to recognize it). To get the list of characters of each movie, we use IMDb.} Finally, each annotation is checked by at least three different people in order to prevent as many errors as possible.
At the end of the process, every track, corresponding to a character appearance in the movie, is associated to his textual mention in the M-VAD captions, if present. \rev{To this end, we manually build a dictionary which maps every character to the set of his names in a movie, and use it for matching textual mentions with tracks. For instance, in the \textit{Robin Hood} movie, Friar Tuck is sometimes referred to as \textit{Tuck} and sometimes as \textit{Friar}; similarly, in \textit{Snow Flower and the Secret Fan}, Nina is sometimes called \textit{Lily}, but also \textit{Flower} and \textit{Sophia}. Once these ambiguities have been solved through the dictionary, each track is mapped to the correct character identity.}

\subsection{M-VAD captions}
Along with the M-VAD Names dataset, we release an extended version of the original M-VAD movie descriptions.
In particular, during the annotation process, we found that several annotated characters were not tagged as ``someone'' in the original M-VAD captions but were mentioned with their proper names. The corresponding captions could be thus considered as errors of the original M-VAD dataset since, as mentioned, existing video captioning architectures are not able to mention a character with its proper name. 

To fix this problem, we add new annotations (\ie new ``someone'' tags) in every movie caption for each mentioned character that is not annotated in the original M-VAD, but that we have correctly annotated in the previous stage of the process. 
Overall, we fix $1,253$ M-VAD descriptions by adding $116$ unique characters that appeared in the original captions but that were not tagged as ``someone''. We remind to Section~\ref{sec:stats} for a comprehensive analysis of the overall statistics of the proposed dataset.

\subsection{Training, validation and test splits}
Original M-VAD training, validation, and test set are obtained by splitting the $92$ movies in three disjoint parts, in order to be able to train video captioning algorithms on a sub-set of movies and to validate and test them on different movies, effectively testing the generalization capabilities of the models.
However, when considering the naming task, video clips of the same movie have to be in every split, so that the captioning algorithms can learn the visual appearance of the characters on the training set and apply it on the validation and test set.
Therefore, we release the official training, validation, and test set for the M-VAD Names dataset.

In particular, we generate the splits applying the following constraints.
Firstly, we forced every movie to have $80\%$ of the video clips into the training set, 10\% into the validation set and $10\%$ into the test set. Secondly, we split the video clips with only one mention, and the video clips with two or more mentions using the same proportions. Finally, we enforced, when possible, to have at least one video clip for every character in each sub-set of the dataset, giving priority to the training set.
Applying this set of soft constraints, training, validation, and test set tend to respectively have $80\%$, $10\%$, and $10\%$ of video clips of each movie, of video clips of each character, of video clips with one mention, and of video clips with two or more mentions.

\setlength{\tabcolsep}{4.5pt}  
\begin{table}[t]
    \caption{Overall statistics of the old and the new version of the M-VAD Names dataset. Along with the number of videos in the train, validation, and test splits, we report the number of mentioned characters, annotated characters, textual mentions, face tracks, and annotated bounding boxes. The numbers reported in the ``Old'' columns refer to statistics reported in~\cite{pini2017towards}.}  
    \label{tab:statistics}
    \renewcommand{\arraystretch}{1.2}
    \centering
    \begin{tabular}{l|C{1.1cm}|C{1.3cm}|C{1.05cm}|C{1.05cm}|C{1.05cm}|C{1.05cm}}                         
     &\multicolumn{2}{c|}{Overall} & \multicolumn{2}{c|}{Avg. per movie} & \multicolumn{2}{c}{Avg. per character} \\ \cline{2-7}
     & Old & New & Old & New & Old & New \\ \hline
        Train videos                            & 17,170 & 19,023     & 187 & 207      & -  & -     \\
        Validation videos                       & 2,708 & 2,976       & 29 & 32        & -  & -     \\ 
        Test videos                             & 2,581 & 2,836       & 28 & 31        & -  & -     \\ \hline
        Mentioned characters                    & 1,450 & 1,566       & 16 & 17        & -  & -     \\
        Annotated characters                    & 908 & 1,093        & 10 & 12        & -  & -     \\
        Mentions                                & 33,073 & 34,388     & 359 & 374      & 23 & 23    \\
        Tracks                                  & 53,665 & 63,442     & 583 & 690      & 62 & 62    \\
        Bounding boxes                & 858,640 & 2,636,595  & 9,328 & 28,658   & 992 & 2,587 \\
    \end{tabular}
\end{table}
\setlength{\tabcolsep}{6pt}

\subsection{Statistics} \label{sec:stats}
In Table~\ref{tab:statistics}, we report the main statistics of the old and the new version of the dataset.
It is worth to notice that the newer version contains more annotated video clips ($24,835$ instead of $22,459$), more unique mentioned characters ($1,566$ instead of $1,450$), and more unique annotated characters ($1,093$ instead of $908$), thanks to the refined track extraction procedure and the extended version of the captions.
With respect to the $34,388$ mentions in the screenplays, the movie characters appear in $63,442$ different face tracks resulting in more than $2$ millions annotated bounding boxes. As it can be noticed, the number of annotated bounding boxes is significantly higher than that of the old version thus confirming the effectiveness of the proposed face detection and tracking procedure. Also, in the previous version, we had tracks with a fixed length of $16$ frames. In the proposed dataset, instead, we have face tracks with a variable length, that on average is equal to $42$.

These statistics refer to the tracks associated with a ``someone'' tag in the caption, while the dataset contains every annotated track, regardless of the existence of the association with a caption tag.
Considering every annotated track, the dataset is composed by more than 100k face tracks and 4M annotated bounding boxes. 
This approach has three main advantages. First, additional annotated tracks can be used for learning the characters' visual appearances since they do not depend on the captions. Then, additional annotated tracks can be linked to caption nouns and pronouns (despite the missing ``someone'' tags) by applying fine NLP or co-reference resolution algorithms. Finally, thanks to the high number of face bounding boxes and their association to specific characters/actors, the dataset can be used for other tasks as well, like action recognition and training of visual-semantic spaces on videos.

\section{Replacing the ``Someone'': an approach to Video Description with Naming}
With the M-VAD Names dataset, we aim to provide sufficient labelled data to allow the development of video captioning architectures with naming capabilities, \ie architectures able to correctly generate captions mentioning proper character names.
Here, we address a strictly related problem that shares many of the challenges of the video captioning with the naming task, yet without considering the generation of movie descriptions.
In particular, we investigate the task of replacing the ``someone'' tags in existing video captions with proper character names. Therefore, we need to analyze both video clips and textual descriptions to find the correct association between visual and textual actions computed by the characters.

\begin{figure}[t]
\centering
\includegraphics[width=\textwidth]{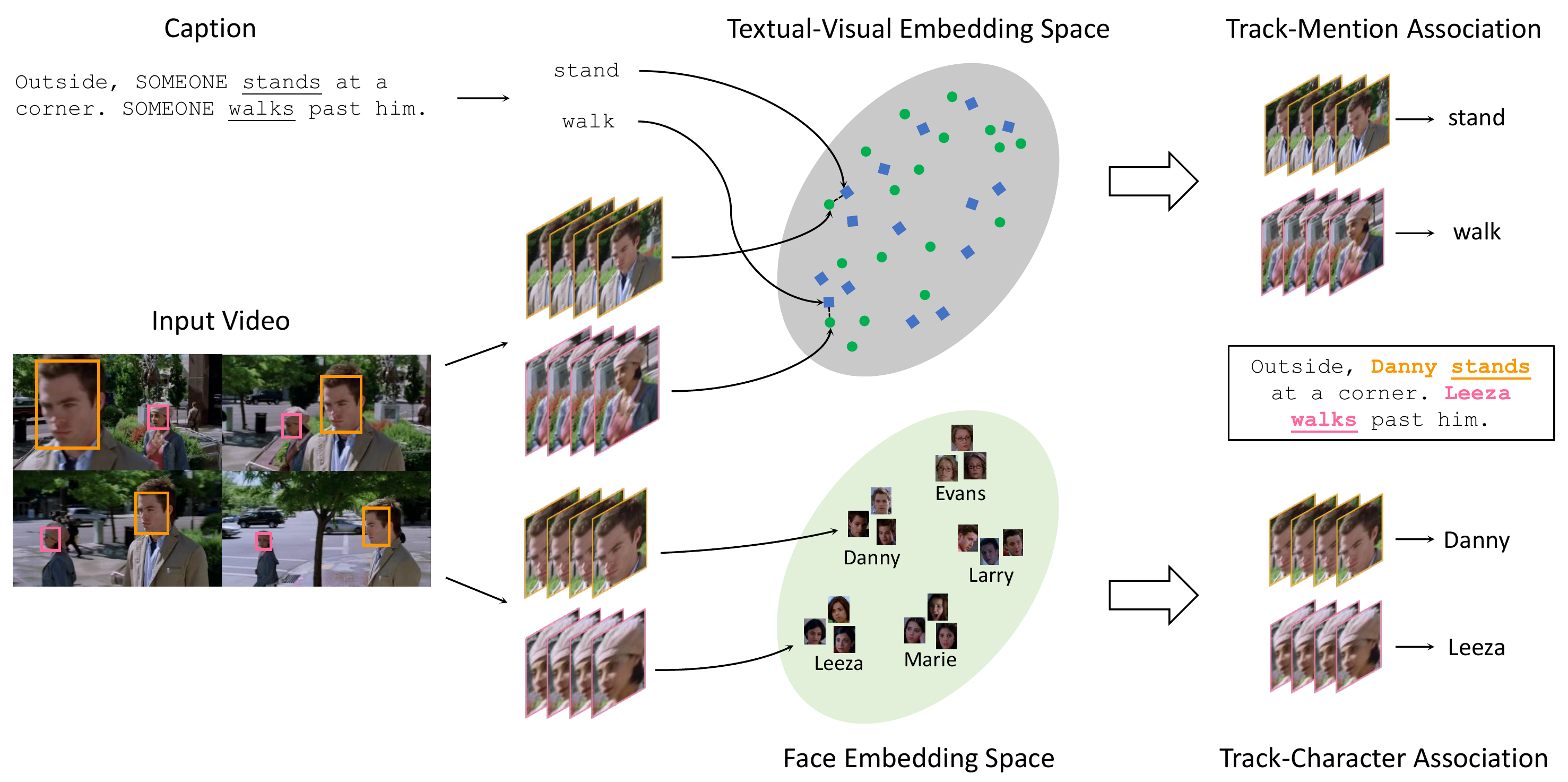}
\caption{Summary of our approach for replacing ``someone'' tags with proper character names. 
We project verbs and body tracks in a unified embedding space to find the best verb-track association. Similarly, we project face tracks in an embedding space to recognize the corresponding characters, completing the replacement task.
}
\label{fig:model}
\end{figure}

A summary of the proposed method is shown in Figure~\ref{fig:model}.
Firstly, we parse ground-truth captions, in which each character name is replaced with a ``someone'' tag, with an NLP parser in order to extract each verb associated to a ``someone'' tag. Then, by using the M-VAD Names dataset, in which characters' visual appearances are associated to their textual mentions, we train a neural network that projects visual actions and textual verbs into a joint multimodal embedding space in which the distance between a verb and a track is inversely proportional to their similarity. After assigning each verb to a visual track, a face recognition algorithm is applied to identify the character and to replace the ``someone'' tag with the correct character name, by concluding the replacement task.

\subsection{Textual-Visual embedding space}\label{sec:mmn}
We project input tracks and verbs into a shared multimodal embedding space in which the distance between a verb and a track is inversely proportional to their similarity.

Regarding the visual data representation, we use the 4096-dimensional output of the last but one fully connected layer of the C3D network~\cite{tran2015learning}, pre-trained on the Sports-1M dataset~\cite{karpathy2014large}, as the visual features for the tracks. In particular, we expand the spatial area of the face tracks to include the upper-body of the subject, as done in~\cite{bojanowski2013finding,pini2017towards}, and we split the tracks in 16-frame long sub-sequences with a stride of 8 frames. Moreover, for each sub-sequence, we fixed the dimension and the position of the track bounding box as the smallest area containing every body bounding box of the sub-sequence. We therefore obtain spatially and temporarily continuous sub-sequences of 16 frames for each original face track. We compute the C3D visual features for each of them. At training time, in order to increase the generalization capabilities of the network, we select a random 16-frame long sub-sequence each time the track is selected, while we average feature vectors of each track, obtaining a single 4096-dimensional vector, at validation and test time.

Regarding the textual data representation, we convert every verb to a 300-dimensional semantic feature vector by using the GloVe embeddings~\cite{pennington2014glove} provided by SpaCy\footnote{\url{https://spacy.io}}, an open-source software library for Natural Language Processing.

Then, we project the visual and the textual features by passing through a fully connected neural network with two branches. The network is trained forcing every track and its corresponding verb to have close projections and forcing every track and every non-corresponding verb to be far by at least a margin $\alpha$.
We can express the cost functions of this formulation as:
\begin{subequations}
    \label{eq:siamese_elems}
    \begin{align}
    p(\mathbf{a},\mathbf{b}) &= \| \phi_v(\mathbf{a}) - \phi_t(\mathbf{b}) \|_2^2, \quad\quad \label{eq:siamese_elems:a} \\
    n(\mathbf{a},\mathbf{b}) &= \max(\alpha - \| \phi_v(\mathbf{a}) - \phi_t(\mathbf{b}) \|_2^2, 0) \label{eq:siamese_elems:b}
    \end{align}
\end{subequations}
or, using a triplet formulation, as:
\begin{subequations}
    \label{eq:triplet_elems}
    \begin{align}
        t(\mathbf{a}, \mathbf{b}, \mathbf{b}^-) &= \max(\| \phi_v(\mathbf{a}) - \phi_t(\mathbf{b}) \|_2^2 - \| \phi_v(\mathbf{a}) - \phi_t(\mathbf{b}^-) \|_2^2 + \alpha, 0) \label{eq:triplet_elems:a} \\
        v(\mathbf{a}, \mathbf{b}, \mathbf{a}^-) &= \max(\| \phi_v(\mathbf{a}) - \phi_t(\mathbf{b}) \|_2^2 - \| \phi_v(\mathbf{a}^-) - \phi_t(\mathbf{b}) \|_2^2 + \alpha, 0) \label{eq:triplet_elems:b}
    \end{align}
\end{subequations}
where $\phi_v(\cdot)$ and $\phi_t(\cdot)$ are respectively the visual-branch and the textual-branch projection function, while $\mathbf{a}$ and $\mathbf{b}$ are the features of a track and of a verb. We denote with $\mathbf{b}^-$ the features of a verb that does not correspond with $\mathbf{a}$ (\ie a verb that is different from $\mathbf{b}$) and with $\mathbf{a}^-$ the features of a track that does not correspond with $\mathbf{b}$ (\ie a track that is not associated to the verb $\mathbf{b}$).

When using the first formulation, a commonly used loss function is the so-called siamese loss, defined as:
\begin{equation}
    \label{eq:siamese}
    L = \sum_{i=1}^N p(\mathbf{a}_i, \mathbf{b}_i) + n(\mathbf{a}_i, \mathbf{b}_i^-)
\end{equation}
where $N$ is the number of valid verb-track pairs.
When using the latter formulation, instead, the so-called triplet loss, or one of its variants, is usually used. The one-term formulation is:
\begin{equation}
    \label{eq:triplet}
    L = \sum_{i=1}^N t(\mathbf{a}_i, \mathbf{b}_i, \mathbf{b}_i^-)
\end{equation}
Recently, a two-term variation has been proposed and successfully employed in~\cite{socher2014grounded,kiros2014unifying,karpathy2015deep,zhu2015aligning}:
\begin{equation}
    \label{eq:two-terms_triplet}
    L = \sum_{i=1}^N t(\mathbf{a}_i, \mathbf{b}_i, \mathbf{b}_i^-) + v(\mathbf{a}_i, \mathbf{b}_i, \mathbf{a}_i^-)
\end{equation}

Addressing our particular task, however, we do not only want to force the proximity of the corresponding tracks and verbs and a minimum distance between the non-corresponding ones, but we also want to force valid verbs and wrong tracks to be far.
In particular, we formulate the following four-terms loss function:
\begin{equation}
    L = \sum_{i=1}^N p(\mathbf{a}_i, \mathbf{b}_i) + n(\mathbf{a}_i, \mathbf{b}_i^-) + p(\mathbf{a}_i^+, \mathbf{b}_i) + n(\mathbf{a}_i^w, \mathbf{b}_i)
    \label{eq:loss}
\end{equation}
where $N$ is the number of valid verb-track pairs, $\mathbf{b}_i^-$ are the features of a verb that does not correspond to the visual track, $\mathbf{a}_i^+$ are the features of a track (different from $\mathbf{a}_i$) associated to the same verb $\mathbf{b}_i$, and $\mathbf{a}_i^w$ are the features of a ``wrong'' track.

Furthermore, since the goal of the whole architecture is to distinguish verb-track pairs extracted from the same video clip, we introduce the following sampling procedure.
Given that we have to select a wrong verb $\mathbf{b}_i^-$ (\ie a verb that is different from the verb $\mathbf{b}_i$) and a positive track $\mathbf{a}_i^+$ (\ie a track, different from $\mathbf{a}_i$, related to the same verb $\mathbf{b}_i$) for each valid verb-track pair, we pick them out from the same video clip of the track $\mathbf{a}_i$ and the verb $\mathbf{b}_i$. If the video clip does not contain $\mathbf{b}_i^-$ or $\mathbf{a}_i^+$, the missing elements are chosen in video clips of the same movie, if possible, otherwise they are randomly chosen between any video clip of the dataset.

Finally, at validation and test time, we compute the distances between verbs and tracks of each video clip and we find the best verb-track association by applying the Kuhn-Munkres algorithm on the distance matrix.

\subsection{Face recognition} \label{sec:face_recognition}
In order to fulfill the replacement of the ``someone'' tags, every track that has been joined to a verb has to be associated to a movie character.
Therefore, we convert each track to a 128-dimensional embedded representation by using a deep neural model inspired by FaceNet~\cite{schroff2015facenet} and trained on a sub-set of the MS-Celeb-1M dataset~\cite{guo2016ms}, as done in Section~\ref{sec:mca}. Then, we classify each embedding using a kd-tree, an optimized version of the K-Nearest Neighbours classifier, fitted on the character embedded representations of the training set.
The K-NN classifier has the advantage of being particularly flexible when considering characters with different visual aspects within the same movie (\ie classes with many clusters lying in different areas of the 128-d embedding space). On the contrary, other classifiers, such as linear models and other types of clustering, are not always capable of correctly classify these cases.

\section{Experimental evaluation}\label{sec:res}
In this section, we report implementation details and experimental results of the proposed architecture for the task of replacing the ``someone'' tags with proper character names.

Along with the proposed loss function, defined in Equation~\ref{eq:loss}, we evaluate the performances of different loss functions. In particular, we test the binary loss function, which is defined as the binary cross entropy on a single label, and the siamese loss function, which is defined in Equation~\ref{eq:siamese}. Moreover, we test the two-terms version of the triplet loss function (Equation~\ref{eq:two-terms_triplet}) and a four-terms variation, which is defined as the two-terms version with the addition of the terms $t(\mathbf{a}_i^+, \mathbf{b}_i, \mathbf{b}_i^-)$ and $v(\mathbf{a}_i, \mathbf{b}_i, \mathbf{a}_i^w)$, where $\mathbf{a}_i^+$, $\mathbf{b}_i^-$, and $\mathbf{a}_i^w$ are defined as in Equation~\ref{eq:loss}.
In addition, we evaluate the siamese, the triplet, and the proposed loss function by using both the euclidean distance and the cosine similarity.

Moreover, we show a comparison of different face recognition methods used to link a face track to a movie character as well as other analyses and qualitative results on the proposed dataset. \rev{After assessing the performance of the proposed architecture on the M-VAD Names dataset, we also test the proposed method in a more general setting, applying it on movies outside of the M-VAD Names. This setting is particularly challenging, as no fine-tuning of the multimodal embedding space is carried out.}

\subsection{Implementation details}
The multimodal neural network that projects textual and visual features into the same embedding space is composed by two branches, formed by one 128-dimensional fully connected layer (with the ReLU activation function) each. The first branch projects the C3D visual features into the embedding space, while the second one projects the GloVe textual features into the same multimodal space. When evaluating the binary loss function, an additional 128-dimensional fully connected layer, which takes as input the concatenation of the two branches with the ReLU activation function, and a one-dimensional fully connected layer with the sigmoid activation function, which predicts the correspondence or non-correspondence of the verb-track pair, are added to the network.

During the training process, we minimize the loss function by applying the Stochastic Gradient Descent using an initial learning rate set to $0.002$ with Nesterov momentum $0.9$ and weight decay $0.0005$.
We use batches composed by 128 random samples. The loss margin $\alpha$ is fixed to $0.2$.
As done in Section~\ref{sec:mca}, we ignore the smallest tracks (\ie face tracks with a side lower than $28$ pixels) in order to prevent the addition of noise during the training phase.

\begin{table}[t]
    \renewcommand{\arraystretch}{1.2}
    \caption{Experimental results on the ``replacing the someone'' task, with different loss functions. Results are reported, in terms of accuracy, on both validation and test splits of the M-VAD Names dataset.}
    \label{tab:results}
    \centering
    \begin{tabular}{L{6.6cm}|C{1.8cm}|C{1.8cm}}
        & Val. Acc. (\%)    & Test Acc. (\%)           \\ \hline
        Random assignment                                       & 11.9     & 11.6 \\  \hline
        Binary with two terms                                   & 48.2     & 49.9 \\
        Triplet Loss with two terms (cosine similarity)                    & 54.3     & 53.6 \\
        Triplet Loss with two terms (euclidean distance)                   & 56.4     & 58.5 \\
        Siamese (cosine similarity)                             & 54.4     & 54.0 \\ 
        Siamese (euclidean distance)                            & 56.1     & 59.0 \\ \hline
        Binary with four terms                                  & 49.8     & 51.0 \\
        Triplet Loss with four terms (cosine similarity)    & 58.7     & 58.2 \\
        Triplet Loss with four terms (euclidean distance)   & 57.8     & \textbf{59.1} \\
        Proposed Loss (cosine similarity)                       & 58.7     & 58.0 \\
        Proposed Loss (euclidean distance)                      & \textbf{60.1}     & 59.0 \\
    \end{tabular}
\end{table}

\subsection{Experimental results}
Table~\ref{tab:results} shows experimental results in terms of the final accuracy of replacing ``someone'' tags in existing captions with proper character names. We report the results of the proposed model trained with all the aforementioned loss functions. For the triplet loss (with two and four terms), the siamese version, and the proposed loss function, results are reported using both the euclidean distance and the cosine similarity. For reference, we also test the results of a random replacement of any ``someone'' tag with a character name randomly extracted from the character list of each movie.

As it can be seen, the proposed strategy of considering positive and negative pairs of verbs and tracks as well as the wrong detections is beneficial for the final accuracy. In particular, on the validation set, the model trained with the proposed loss obtains the best performances. On the testing set, instead, the model trained with the triplet loss with four terms is the best performing one, even though by a slight margin. 

\begin{figure}[t]
    \centering
    \subfloat[Proposed Loss (cosine similarity)]{
        \includegraphics[width=.43\textwidth]{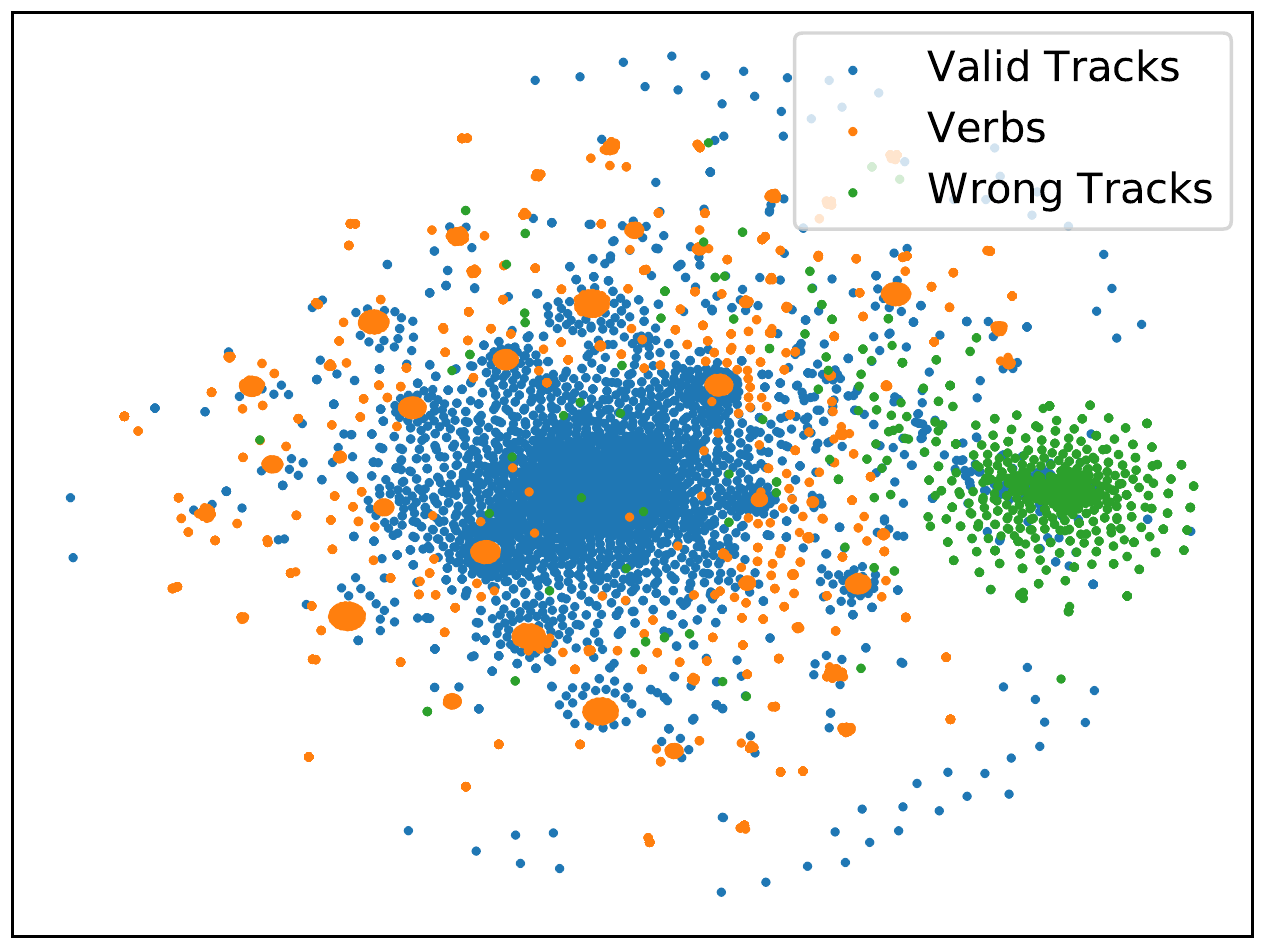}
    }\quad
    \subfloat[\footnotesize{Proposed Loss (euclidean distance)}]{
        \includegraphics[width=.43\textwidth]{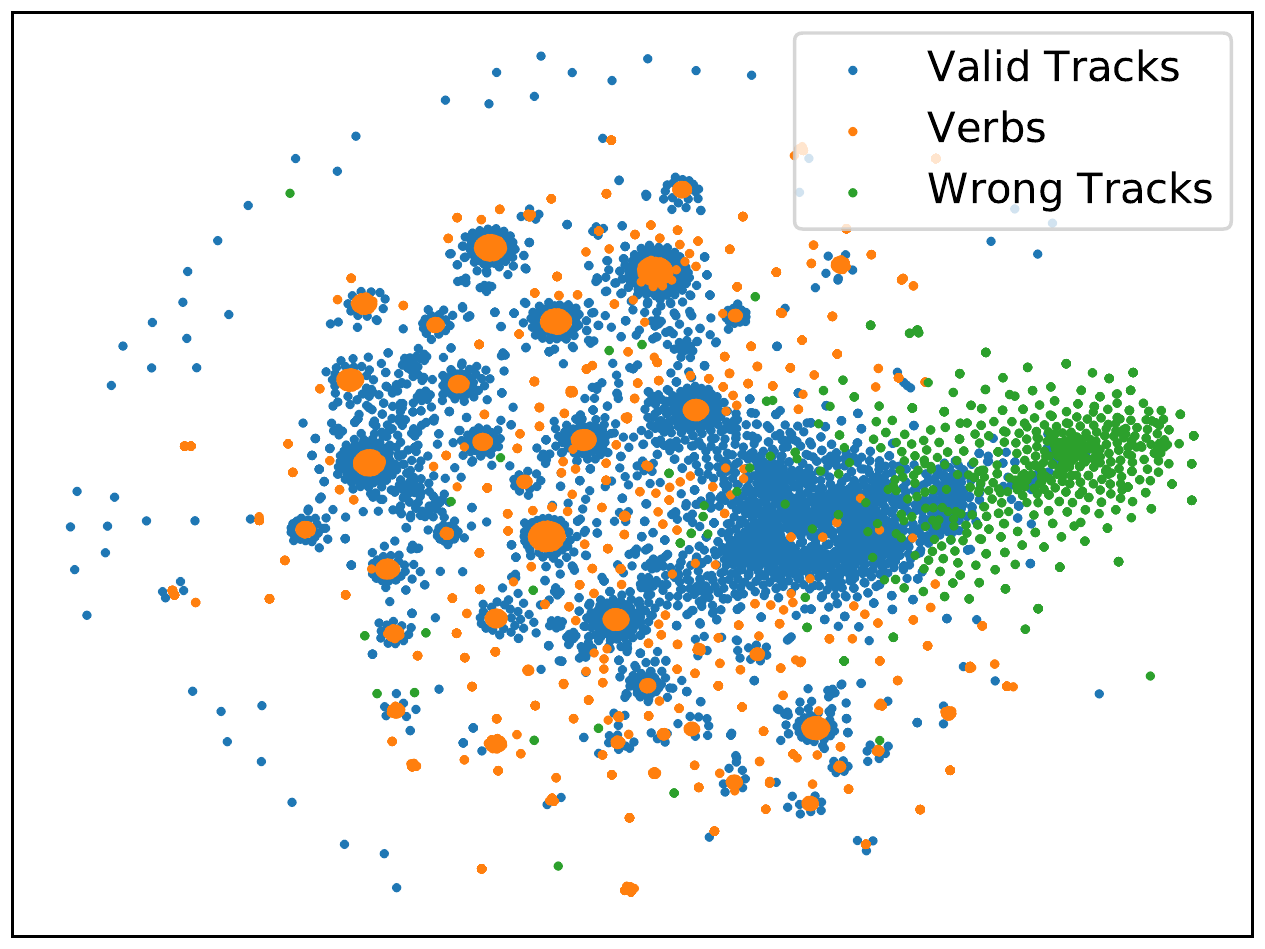}
    }
    \caption{Comparison between textual-visual embedding spaces obtained by training the model with the proposed loss function using both euclidean distance and cosine similarity. Visualization is obtained by running the t-SNE algorithm on top of the verb-track embedded representations. Best seen in color.}
    \label{fig:tsne}
\end{figure}

Figure~\ref{fig:tsne} shows a representation of the textual-visual embedding space obtained by training the model with the proposed loss function using both the euclidean distance and the cosine similarity. In particular, we report each verb-track pair of the M-VAD Names validation set along with all wrong visual tracks of the corresponding video clips. To get a suitable two-dimensional representation out of a 128-dimensional space, we run the t-SNE algorithm~\cite{maaten2008visualizing,van2014accelerating}, which iteratively finds a non-linear projection which preserves pairwise distances from the original space. As it can be noticed, the represented embedding spaces are composed by clusters of verb representations that probably correspond to verbs with a similar meaning. Wrong tracks are discriminated quite well in both spaces, while valid tracks are better divided and assigned to a specific verb cluster when using the euclidean distance thus confirming the quantitative results reported in Table~\ref{tab:results}. 

\begin{figure}[t]
    \centering
    \includegraphics[width=.8\textwidth]{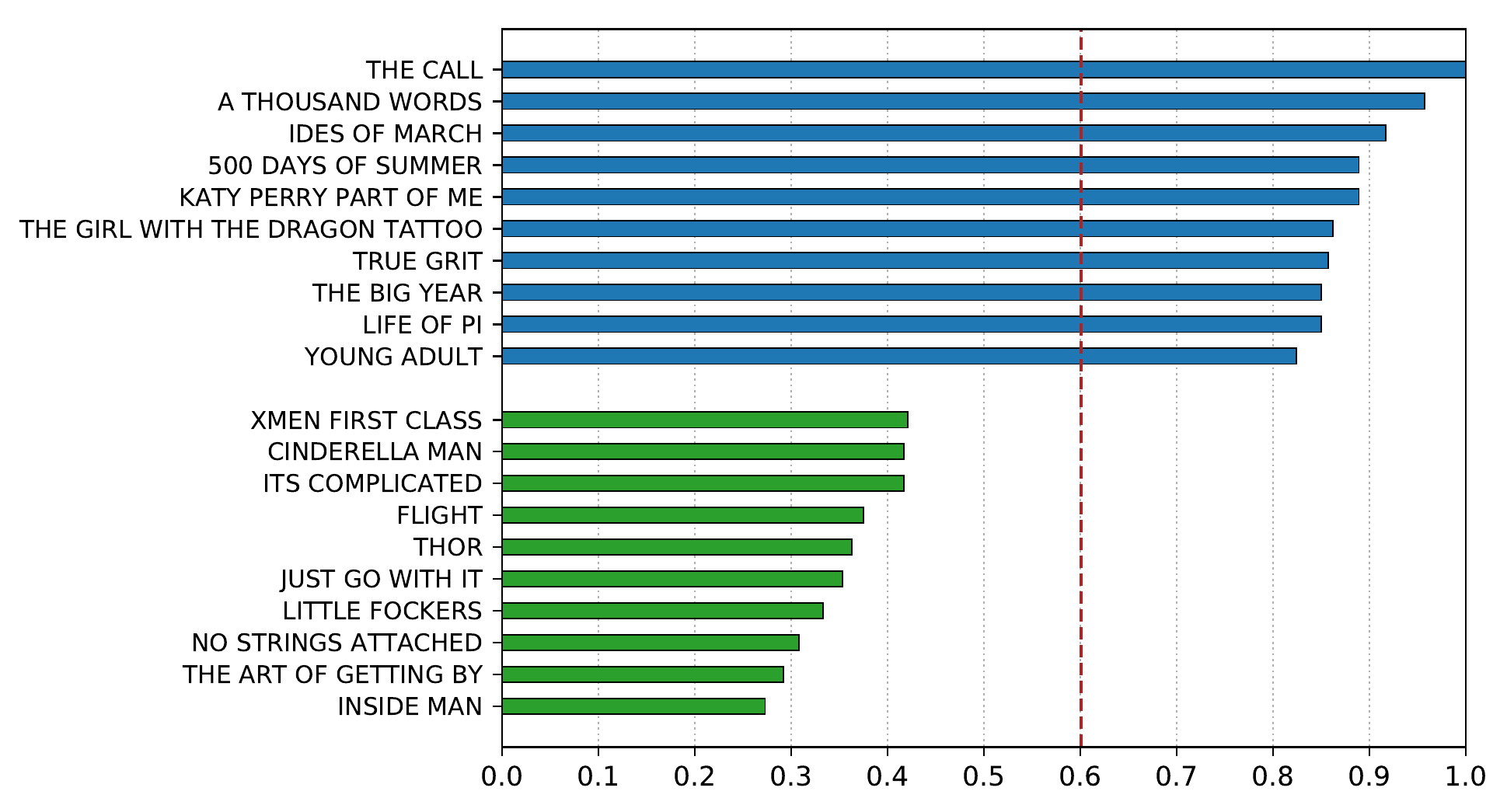}
    \caption{Accuracy results on single movies of the proposed approach on the M-VAD Names validation set. We report the $10$ best results and $10$ worst ones, respectively represented in blue and green. The vertical line represents the averaged accuracy on all movies.}
    \label{fig:hist_film}
\end{figure}

In Figure~\ref{fig:hist_film}, we also report the results in terms of validation accuracy obtained on single movies. In particular, the graph shows the results obtained on the $10$ movies with the best accuracy results and those obtained on the $10$ movies with the worst ones. These results highlight that correct verb-track matches are more difficult for a specific sub-set of movies. This is probably due to different number of characters or different number of unknown and wrong tracks that could cause greater difficulty in associating a verb with its corresponding visual track.

To validate the face recognition method, in Table~\ref{tab:facerec_comparison} we report a comparison between different classifiers. In details, we compare the K-Nearest Neighbours (K-NN) classifier (with a $k$ value of $5$) with the SVM (applying the Radial Basis Function kernel) and the Adaboost (with $30$ Decision Trees) classifier. For each of them, we only use the face tracks of the training set to classify all validation and test samples, as previously mentioned in Section~\ref{sec:face_recognition}. As it can be seen, the K-NN performs better on both the validation and the test set.

In order to assess the effectiveness of the proposed sampling strategy, described in Section~\ref{sec:mmn}, we compare it by sampling the verb-track pairs within the whole dataset. Table~\ref{tab:sampling_results} shows the validation accuracies of the different sampling strategies. Results confirm that the proposed procedure of sampling verb-track pairs within the same video clip, if possible, allows to better discriminate samples from the same movie.

\begin{table}[t]
    \renewcommand{\arraystretch}{1.2}
    \caption{Accuracy of different character face classifiers. Accuracy is calculated as the number of correct predictions on the known-character tracks of the validation and the test set.}
    \label{tab:facerec_comparison}
    \centering
    \begin{tabular}{L{2.56cm}|C{1.8cm}|C{1.8cm}}              
                    & Val. Acc. (\%)    & Test Acc. (\%)          \\ \hline
        K-NN        & \textbf{85.2}    & \textbf{86.3}    \\  
        SVM         & 84.2             & 85.2             \\  
        Adaboost    & 64.7             & 65.8             \\
    \end{tabular}
\end{table}

\begin{table}[t]
    \renewcommand{\arraystretch}{1.2}
    \caption{Comparison of different sampling strategies on the M-VAD Names validation set using the proposed loss function to train the model.}
    \label{tab:sampling_results}
    \centering
    \begin{tabular}{L{4.8cm}|C{1.8cm}}
                                    & Val. Acc. (\%)         \\ \hline
        Sampling within a video clip         & \textbf{60.1}     \\
        Sampling within the whole dataset         & 58.8              \\
    \end{tabular}
\end{table}

\subsection{Generalization capabilities of the proposed approach}
\rev{After assessing the performance of the proposed architecture on the M-VAD Names dataset, in which the training, the validation, and the test split share the same character names, we address a more challenging and realistic evaluation scenario. In this case, we evaluate on movies outside of the M-VAD Names, while using the multimodal embedding space trained on the proposed dataset. To link characters' appearances with their identities, we initialize the face embedding space by randomly sampling 10\% of the face tracks. Beyond this limited supervision signal, which is mandatory when new characters are added, we do not exploit any other training data related to the new set of movies.}

\rev{The set of external videos contains three movies belonging to the MPII-MD dataset for video captioning~\cite{rohrbach15cvpr}, namely \textit{Harry Potter and the Philosopher’s Stone}, \textit{Pulp Fiction} and \textit{Sherlock Holmes: A Game of Shadows}. Table~\ref{tab:mpii} shows the accuracy results for both the face classification and the ``someone'' replacement task. Numbers are reported on the portion of tracks which were not used for the initialization of the face embedding space. The overall accuracy is reported by averaging on every considered track (notice that each movie has a different number of tracks). As it can be noticed, the accuracy of the replacement task is similar to the one obtained when testing on the M-VAD Names, thus confirming the generalization capabilities of the proposed model.}

\subsection{Qualitative results}
\begin{figure}[t]
\centering
    \includegraphics[width=\textwidth]{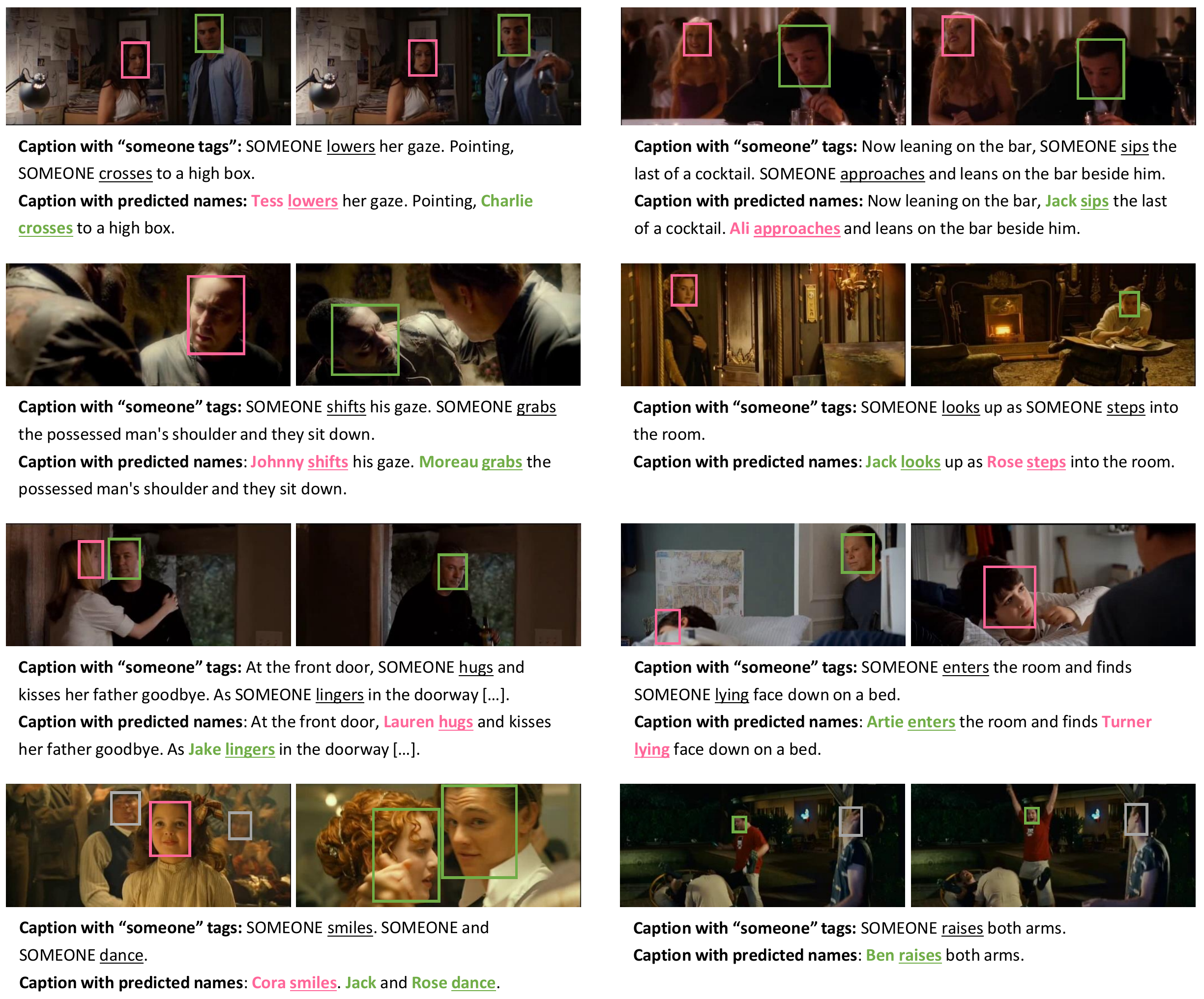}  
\caption{Sample results of the proposed method for replacing ``someone'' tags with character proper names. For each sample, tracks associated with the same verb are represented with the same color, while tracks that are not associated with any verb are reported in gray.}
\label{fig:results}
\end{figure}

Figure~\ref{fig:results} shows some qualitative results on sample clips from the M-VAD Names validation set. For each movie clip, we report the original caption with ``someone'' tags and that with the corresponding character names predicted by our approach. As it can be seen, our model is able to discriminate tracks containing different actions and to associate them with the corresponding verb in the captions. Also, visual tracks of unknown characters or character tracks that are not associated to a verb in the caption are correctly not paired to any verb, as for example in the fourth row of the figure.

\begin{table}[t!]
    \renewcommand{\arraystretch}{1.2}
    \caption{\rev{Performance on an external set of movies from the MPII-MD dataset~\cite{rohrbach15cvpr}, using 10\% of the tracks for training the face embedding space, and the multimodal embedding space model pre-trained on M-VAD Names.}}
    \label{tab:mpii}
    \centering
    \begin{tabular}{L{5.6cm}|C{2.3cm}|C{2.3cm}}              
                                                    & Face Class. (\%)  & Replacement (\%)    \\ \hline
        Harry Potter and the Philosopher's Stone    & 78.6              & 59.0              \\
        Pulp Fiction                                & 81.4              & 54.6              \\
        Sherlock Holmes: A Game of Shadows          & 82.0              & 65.6              \\ \hline
        Overall accuracy                   & \textbf{80.7}     & \textbf{60.1}     \\
    \end{tabular}
\end{table}

\begin{figure}[t]
\centering
    \includegraphics[width=\textwidth]{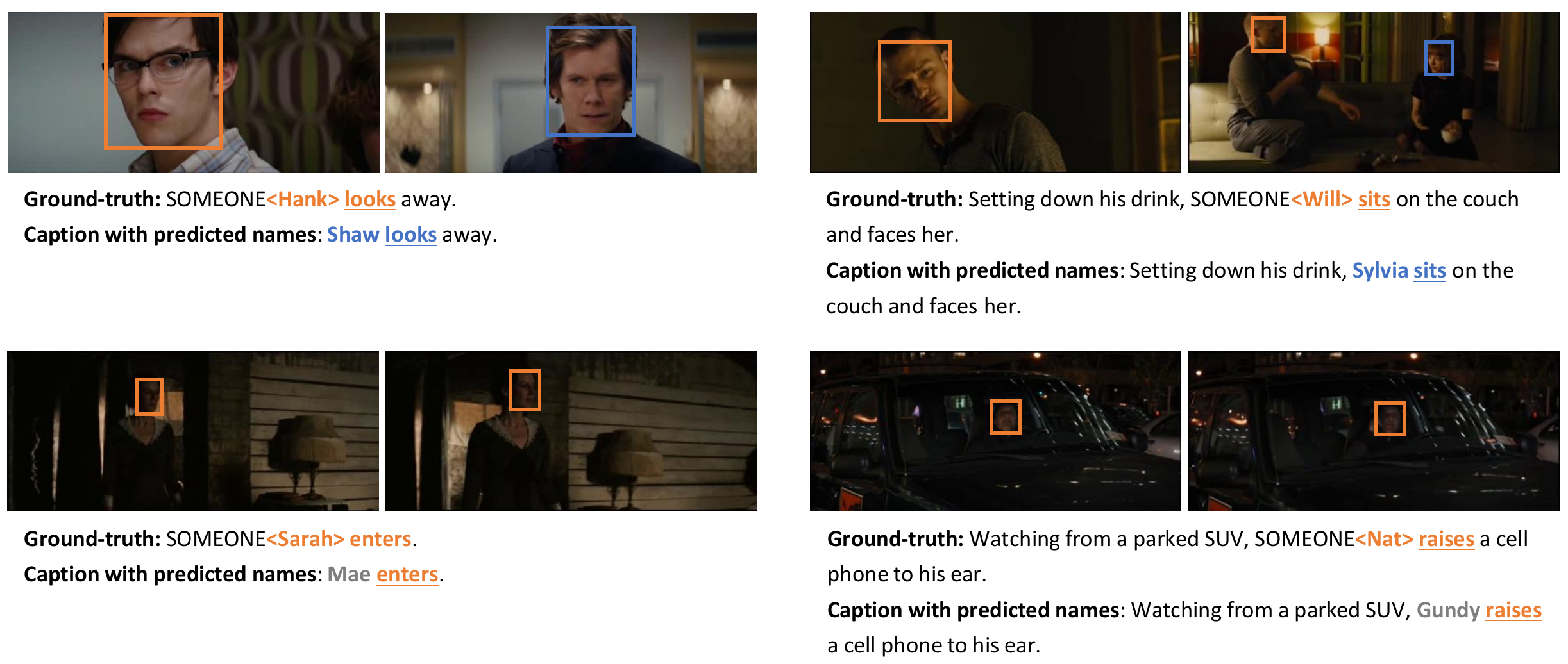}  
\caption{Failure cases from the M-VAD validation set. In the first row, two track-verb association errors, while in the second row, two face recognition errors.}
\label{fig:failures}
\end{figure}

Finally, we report some failure cases in Figure~\ref{fig:failures}. In particular, the figure shows two verb-track association errors (first row), and two cases in which the error is due to the face recognition phase (second row). In the first case, the verb in the caption is associated with a different visual track of the considered movie clip that is then correctly classified with the corresponding character name. In the other one, instead, the visual track is correctly associated with the corresponding verb in the caption, but the face recognition algorithm fails to identify the correct character appearing in the movie clip.

\section{Conclusion}
In this paper, we introduced a novel version of the M-VAD Names dataset, specifically designed for supporting the development of video captioning architectures with naming capabilities. The dataset, which is an extension of the well-known M-VAD dataset, consists of visual face tracks and their association with characters' textual mentions.
Moreover, we presented a multimodal architecture that addresses the task of replacing generic ``someone'' tags with proper character names in previously generated captions. The model combines advanced Natural Language Processing tools and state-of-the-art deep neural models for action and face recognition.
Experimental results demonstrated, through extensive analyses on the proposed dataset, the effectiveness of the devised solutions and highlighted the challenges of the considered task.

\begin{acknowledgements}
We acknowledge Carmen Sabia and Luca Bergamini for supporting us during the annotation of the M-VAD Names dataset.
We also gratefully acknowledge Facebook AI Research and Panasonic Corporation for the donation of the GPUs used in this work.
\end{acknowledgements}

\bibliographystyle{spmpsci}      
\bibliography{bibliography}   

\begin{thebibliography}{10}
\providecommand{\url}[1]{{#1}}
\providecommand{\urlprefix}{URL }
\expandafter\ifx\csname urlstyle\endcsname\relax
  \providecommand{\doi}[1]{DOI~\discretionary{}{}{}#1}\else
  \providecommand{\doi}{DOI~\discretionary{}{}{}\begingroup
  \urlstyle{rm}\Url}\fi

\bibitem{Babenko09visualtracking}
Babenko, B., Yang, M.H., Belongie, S.: Visual tracking with online multiple
  instance learning.
\newblock In: IEEE International Conference on Computer Vision and Pattern
  Recognition (2009)

\bibitem{baraldi2017hierarchical}
Baraldi, L., Grana, C., Cucchiara, R.: Hierarchical boundary-aware neural
  encoder for video captioning.
\newblock In: IEEE International Conference on Computer Vision and Pattern
  Recognition (2017)

\bibitem{bojanowski2013finding}
Bojanowski, P., Bach, F., Laptev, I., Ponce, J., Schmid, C., Sivic, J.: Finding
  actors and actions in movies.
\newblock In: IEEE International Conference on Computer Vision (2013)

\bibitem{ding2010learning}
Ding, L., Yilmaz, A.: Learning relations among movie characters: A social
  network perspective.
\newblock In: European Conference on Computer Vision (2010)

\bibitem{donahue2015long}
Donahue, J., Anne~Hendricks, L., Guadarrama, S., Rohrbach, M., Venugopalan, S.,
  Saenko, K., Darrell, T.: Long-term recurrent convolutional networks for
  visual recognition and description.
\newblock In: IEEE International Conference on Computer Vision and Pattern
  Recognition (2015)

\bibitem{everingham2006hello}
Everingham, M., Sivic, J., Zisserman, A.: {``Hello! My name is...
  Buffy''--Automatic Naming of Characters in TV Video.}
\newblock In: British Machine Vision Conference (2006)

\bibitem{guo2016ms}
Guo, Y., Zhang, L., Hu, Y., He, X., Gao, J.: {MS-Celeb-1M: A dataset and
  benchmark for large-scale face recognition}.
\newblock In: European Conference on Computer Vision (2016)

\bibitem{hendricks2015deep}
Hendricks, L.A., Venugopalan, S., Rohrbach, M., Mooney, R., Saenko, K.,
  Darrell, T.: Deep compositional captioning: Describing novel object
  categories without paired training data.
\newblock In: IEEE International Conference on Computer Vision and Pattern
  Recognition (2016)

\bibitem{jin2017end}
Jin, S., Su, H., Stauffer, C., Learned-Miller, E.: {End-to-end Face Detection
  and Cast Grouping in Movies Using Erdos-R{\'e}nyi Clustering}.
\newblock In: IEEE International Conference on Computer Vision (2017)

\bibitem{karpathy2015deep}
Karpathy, A., Fei-Fei, L.: Deep visual-semantic alignments for generating image
  descriptions.
\newblock In: IEEE International Conference on Computer Vision and Pattern
  Recognition (2015)

\bibitem{karpathy2014large}
Karpathy, A., Toderici, G., Shetty, S., Leung, T., Sukthankar, R., Fei-Fei, L.:
  Large-scale video classification with convolutional neural networks.
\newblock In: IEEE International Conference on Computer Vision and Pattern
  Recognition (2014)

\bibitem{kiros2014unifying}
Kiros, R., Salakhutdinov, R., Zemel, R.S.: Unifying visual-semantic embeddings
  with multimodal neural language models.
\newblock arXiv preprint arXiv:1411.2539  (2014)

\bibitem{krishna2017dense}
Krishna, R., Hata, K., Ren, F., Fei-Fei, L., Niebles, J.C.: {Dense-Captioning
  Events in Videos}.
\newblock In: IEEE International Conference on Computer Vision (2017)

\bibitem{kuhn1955hungarian}
Kuhn, H.W.: The hungarian method for the assignment problem.
\newblock Naval research logistics quarterly \textbf{2}(1-2), 83--97 (1955)

\bibitem{maaten2008visualizing}
Maaten, L.v.d., Hinton, G.: {Visualizing data using t-SNE}.
\newblock Journal of Machine Learning Research \textbf{9}, 2579--2605 (2008)

\bibitem{marin2014detecting}
Mar{\'\i}n-Jim{\'e}nez, M.J., Zisserman, A., Eichner, M., Ferrari, V.:
  Detecting people looking at each other in videos.
\newblock International Journal of Computer Vision \textbf{106}(3), 282--296
  (2014)

\bibitem{miech2017learning}
Miech, A., Alayrac, J.B., Bojanowski, P., Laptev, I., Sivic, J.: {Learning from
  video and text via large-scale discriminative clustering}.
\newblock In: IEEE International Conference on Computer Vision (2017)

\bibitem{pan2016hierarchical}
Pan, P., Xu, Z., Yang, Y., Wu, F., Zhuang, Y.: Hierarchical recurrent neural
  encoder for video representation with application to captioning.
\newblock In: IEEE International Conference on Computer Vision and Pattern
  Recognition (2016)

\bibitem{pan2015jointly}
Pan, Y., Mei, T., Yao, T., Li, H., Rui, Y.: Jointly modeling embedding and
  translation to bridge video and language.
\newblock IEEE International Conference on Computer Vision and Pattern
  Recognition  (2016)

\bibitem{patron2012structured}
Patron-Perez, A., Marszalek, M., Reid, I., Zisserman, A.: {Structured learning
  of human interactions in TV shows}.
\newblock IEEE Transactions on Pattern Analysis and Machine Intelligence
  \textbf{34}(12), 2441--2453 (2012)

\bibitem{pennington2014glove}
Pennington, J., Socher, R., Manning, C.D.: {GloVe: Global Vectors for Word
  Representation.}
\newblock In: Conference on Empirical Methods in Natural Language Processing
  (2014)

\bibitem{pini2017towards}
Pini, S., Cornia, M., Baraldi, L., Cucchiara, R.: Towards video captioning with
  naming: a novel dataset and a multi-modal approach.
\newblock In: International Conference on Image Analysis and Processing (2017)

\bibitem{ramanathan2014linking}
Ramanathan, V., Joulin, A., Liang, P., Fei-Fei, L.: Linking people in videos
  with ``their'' names using coreference resolution.
\newblock In: European Conference on Computer Vision (2014)

\bibitem{rohrbach2015long}
Rohrbach, A., Rohrbach, M., Schiele, B.: The long-short story of movie
  description.
\newblock In: German Conference on Pattern Recognition (2015)

\bibitem{rohrbach15cvpr}
Rohrbach, A., Rohrbach, M., Tandon, N., Schiele, B.: A dataset for movie
  description.
\newblock In: IEEE International Conference on Computer Vision and Pattern
  Recognition (2015)

\bibitem{RohrbachCVPR2017a}
Rohrbach, A., Rohrbach, M., Tang, S., Oh, S.J., Schiele, B.: {Generating
  Descriptions with Grounded and Co-Referenced People}.
\newblock In: IEEE International Conference on Computer Vision and Pattern
  Recognition (2017)

\bibitem{schroff2015facenet}
Schroff, F., Kalenichenko, D., Philbin, J.: Facenet: A unified embedding for
  face recognition and clustering.
\newblock In: IEEE International Conference on Computer Vision and Pattern
  Recognition (2015)

\bibitem{shetty2017speaking}
Shetty, R., Rohrbach, M., Hendricks, L.A., Fritz, M., Schiele, B.: {Speaking
  the Same Language: Matching Machine to Human Captions by Adversarial
  Training}.
\newblock In: IEEE International Conference on Computer Vision (2017)

\bibitem{sivic2009you}
Sivic, J., Everingham, M., Zisserman, A.: {``Who are you?'' Learning person
  specific classifiers from video}.
\newblock In: IEEE International Conference on Computer Vision and Pattern
  Recognition (2009)

\bibitem{socher2014grounded}
Socher, R., Karpathy, A., Le, Q.V., Manning, C.D., Ng, A.Y.: Grounded
  compositional semantics for finding and describing images with sentences.
\newblock Transactions of the Association of Computational Linguistics
  \textbf{2}(1), 207--218 (2014)

\bibitem{tapaswi2012knock}
Tapaswi, M., B{\"a}uml, M., Stiefelhagen, R.: {``Knock! Knock! Who is it?''
  probabilistic person identification in TV-series}.
\newblock In: IEEE International Conference on Computer Vision and Pattern
  Recognition (2012)

\bibitem{torabi2015using}
Torabi, A., Pal, C., Larochelle, H., Courville, A.: Using descriptive video
  services to create a large data source for video annotation research.
\newblock arXiv preprint arXiv:1503.01070  (2015)

\bibitem{tran2015learning}
Tran, D., Bourdev, L., Fergus, R., Torresani, L., Paluri, M.: Learning
  spatiotemporal features with 3d convolutional networks.
\newblock In: IEEE International Conference on Computer Vision (2015)

\bibitem{van2014accelerating}
Van Der~Maaten, L.: {Accelerating t-SNE using tree-based algorithms}.
\newblock Journal of Machine Learning Research \textbf{15}(1), 3221--3245
  (2014)

\bibitem{venugopalan16emnlp}
Venugopalan, S., Hendricks, L.A., Mooney, R., Saenko, K.: Improving lstm-based
  video description with linguistic knowledge mined from text.
\newblock In: Conf. on Empirical Methods in Natural Language Processing (2016)

\bibitem{venugopalan2015sequence}
Venugopalan, S., Rohrbach, M., Donahue, J., Mooney, R., Darrell, T., Saenko,
  K.: Sequence to sequence-video to text.
\newblock In: IEEE International Conference on Computer Vision (2015)

\bibitem{venugopalan2014translating}
Venugopalan, S., Xu, H., Donahue, J., Rohrbach, M., Mooney, R., Saenko, K.:
  Translating videos to natural language using deep recurrent neural networks.
\newblock North American Chapter of the Association for Computational
  Linguistics  (2014)

\bibitem{moviegraphs2017}
Vicol, P., Tapaswi, M., Castrejon, L., Fidler, S.: {MovieGraphs: Towards
  Understanding Human-Centric Situations from Videos}.
\newblock In: IEEE International Conference on Computer Vision and Pattern
  Recognition (2018)

\bibitem{ward1963hierarchical}
Ward, J.H.J.: Hierarchical grouping to optimize an objective function.
\newblock Journal of the American statistical association \textbf{58}(301),
  236--244 (1963)

\bibitem{yao2015describing}
Yao, L., Torabi, A., Cho, K., Ballas, N., Pal, C., Larochelle, H., Courville,
  A.: Describing videos by exploiting temporal structure.
\newblock In: IEEE International Conference on Computer Vision (2015)

\bibitem{yu2016video}
Yu, H., Wang, J., Huang, Z., Yang, Y., Xu, W.: Video paragraph captioning using
  hierarchical recurrent neural networks.
\newblock In: IEEE International Conference on Computer Vision and Pattern
  Recognition (2016)

\bibitem{ZhangZL016}
Zhang, K., Zhang, Z., Li, Z., Qiao, Y.: Joint face detection and alignment
  using multitask cascaded convolutional networks.
\newblock IEEE Signal Processing Letters \textbf{23}(10), 1499--1503 (2016)

\bibitem{zhu2015aligning}
Zhu, Y., Kiros, R., Zemel, R., Salakhutdinov, R., Urtasun, R., Torralba, A.,
  Fidler, S.: Aligning books and movies: Towards story-like visual explanations
  by watching movies and reading books.
\newblock In: IEEE International Conference on Computer Vision and Pattern
  Recognition (2015)

\end{thebibliography}

\end{document}